# SCK: A SPARSE CODING BASED KEY-POINT DETECTOR

*Thanh Hong-Phuoc, Yifeng He, Ling Guan*

Ryerson Multimedia Research Laboratory, Ryerson University, Toronto, Canada M5B 2K3

## ABSTRACT

All current popular hand-crafted key-point detectors such as Harris corner, MSER, SIFT, SURF… rely on some specific pre-designed structures for the detection of corners, blobs, or junctions in an image. In this paper, a novel sparse coding based key-point detector which requires no particular pre-designed structures is presented. The key-point detector is based on measuring the complexity level of each block in an image to decide where a key-point should be. The complexity level of a block is defined as the total number of non-zero components of a sparse representation of that block. Generally, a block constructed with more components is more complex and has greater potential to be a good key-point. Experimental results on Webcam and EF datasets [1, 2] show that the proposed detector achieves significantly high repeatability compared to hand-crafted features, and even outperforms the matching scores of the state-of-the-art learning based detector.

*Index Terms*— Key-point, interest point, feature detector, sparse coding, sparse representation

## 1. INTRODUCTION

Detection of key-points has been researched for decades, and a large number of successful key-point detectors have been proposed, which provide a foundation for a vast number of applications in computer vision such as image matching, image registration and object tracking. Currently, there are two main categories of key-point detectors: hand-crafted detectors and learning based detectors.

Hand-crafted detectors have attracted much attention since the introduction of corner detection works by Forstner et al., and Harris et al. in the 1980s [3, 4]. Several other outstanding hand-crafted detectors are SIFT [5], SURF [6], MSER [7], and SFOP [8]. Although hand-crafted detectors generally produce good results, they often rely on specific pre-designed structures. In fact, hand-crafted features are known not to be very flexible in different contexts [1, 9] probably due to the nature of pre-designed structures. In practice, there are situations in which a large number of high quality key-points are expected, but cannot be easily located by one specific detector. A common solution is a careful re-selection of an alternative detector or combination of multiple. However, in such cases, if multiple structure types could be detected with a single detector, it would be a simple but very effective solution.

Learning based detectors have received increased focus with the recent success of deep learning. The typical examples of learning based detectors are TIDLE [1] and its following work [9]. These works demonstrate significant improvements in the repeatability of the detected key-points in many challenging datasets. However, to achieve this high performance, the state-of-the-art learning based feature detector [9] relies on the output of TIDLE which was trained on a dataset originating from the results of SIFT detectors. Thus, the development of hand-crafted detectors or other type of detectors for creating reliable inputs for the machine learning algorithms is still an important task.

In this paper, we introduce a key-point detector which requires no particular pre-designed structures. The proposed detector is shown to be able to detect all types of structures. The detector outperforms the existing hand-crafted feature detectors in terms of the repeatability metric developed by Mikolajczyk et al. [10] on Webcam and EF datasets [1, 2]. Experimental results show that the proposed detector even works better than the current top learning based detector on the matching score metric [10]. The proposed key-point detector is based on counting the total number of non-zero components in the sparse representation of each block in an image. We define the number of non-zero components in a sparse representation as the complexity measure. A formal mathematical definition of the measure is presented in Section 3.2. In general, the more components a block requires to construct, the more complex the block is. Any blocks with the complexity measures falling in a specified range can be considered as good key-points. In the event, the number of expected key-points is limited, suitable key-points can be selected based on a proposed novel strength measure. Our contributions are summarized below:

1. We propose a novel sparse coding based key-point detector requiring no particular hand-crafted structure.
2. We introduce a measure to compare the strength of any key-points detected by the proposed detector.

The rest of the paper is organized as follows. In Section 2, we review some popular hand crafted features, and two top learning based feature detectors. In Section 3, the proposed key-point detector is presented. The evaluation results of the proposed detector are reported in Section 4, and the conclusion is presented in Section 5.

## 2. RELATED WORK

### 2.1. Hand-crafted feature detector

The development of hand-crafted detectors often begins with a human designed structure such as a corner, blob, or junction with some specially predefined characteristics such that the structures are stable and easily found in different images under various transformations. Some typical examples of the detectors of this type are Harris corners [4] for corner detection, SIFT [5], SURF [6], MSER [7] for blob detection, and SFOP [8] for junction detection. Besides the list of the aforementioned detectors, there are a vast number of detectors such as SIFER [11], D-SIFER [12], WADE [13], Edge Foci [2] targeting detection of different structures with various customizations. Although current detectors rely on some more or less different pre-designed structures, the structures share a common factor in that they have some levels of complexity. If we can measure the complexity of the structures of all blocks in an image, we do not need a pre-defined structure for a key-point detector, which is the motivation for our work.

### 2.2. Learning based feature detector

Recently, a new trend has emerged regarding the enhanced repeatability of key-point based learning methods. Two of the most prominent works are TIDLE [1] and the work in [9] which is developed based on TIDLE's output. In TIDLE, the key-points in overlapped images captured at different times and seasons with the same viewpoints are first detected by SIFT detectors. Then only blocks around key-points with sufficiently high repeatability are selected as positive samples to train their systems. In [9], the outputs of TIDLE are used as standard patches to train their systems. Both works deliver outstanding improvements in repeatability compared to purely hand-crafted features. However, the works more or less rely on the selection of hand-crafted features to build the training datasets. In other words, a good method to detect reliable key-points as inputs to the training systems is still necessary. The learning based feature detectors, on the other hand, are good ways to enhance the performance in terms of higher repeatability or faster computation.

## 3. SPARSE CODING BASED KEY-POINT DETECTOR

Sparse coding has been widely applied in many different disciplines such as machine learning, neuroscience, and signal processing [14, 15]. The idea of sparse coding is to represent an input signal as the sparsest combination of atoms in a given dictionary. In this section, the proposed sparse coding based algorithm for the detection of key-points in a gray-scale image is presented as follows:

### 3.1. Pre-filtering

A small size low pass filter such as a Gaussian filter is applied to remove noise and small structures in the image.

### 3.2. Sparse coding based key-point detection

A loop through every block $X$ of size $n$ by $n$ in the image is created. For each block, after reshaping $X$ to the size $n^2$ by 1, the mean is subtracted from $X$. Then the difference is normalized to unit amplitude which is noted as $X'$. Next, the sparse representation of the block is found by solving the following optimization problem:

$$\boldsymbol{\alpha} = \mathrm{argmin}(\|X' - D\boldsymbol{\alpha}\| + \lambda\|\boldsymbol{\alpha}\|_{l1}) \quad (1)$$

where $\boldsymbol{\alpha}$ is the sparse representation of the block, $D$ is a complete or over-complete dictionary of the sparse coding. The size of the dictionary is $n^2$ by $k$ where $k$ is the number of atoms in the dictionary. The size of $\boldsymbol{\alpha}$ is $k$ by 1.

The complexity measure ($CM$) is then defined as follows:

$$CM = \|\boldsymbol{\alpha}\|_{l0} \quad (2)$$

where $\|\boldsymbol{\alpha}\|_{l0}$ is the total number of non-zero components in the sparse representation.

If the $CM$ of a block falls into a range specified between a lower and upper limit, the point at the center of the block is considered as a key-point. The lower limit is used to guarantee that the detected key-point has a high enough complexity level to be distinct from other points in the image. The upper limit is used to avoid the detection of overly distinctive key-points, which may not be repeatable in different images of the same scenes. The size of the key-point is then selected as the square root of 2 multiplied by half of the block size to cover all pixels (all possible positions of the key-point) belonging to the block. For example, if a center of a block of size 9x9 is considered as a key-point, the size of the key-point is $4.5\sqrt{2}$. The key-point and size selection techniques are application configurable parameters, which provide flexibility. Depending on application, the entire block may be considered as a key-point, or the size of the key-point may be defined in difference ways.

Since biological evidence has revealed that sparse representations are similar to the receptive fields of simple cells in the primary visual cortex [16], sparse coding has been extensively applied in many saliency detectors such as [16, 17], which look similar to the presented detection method at the first glance. However, the saliency detection works are different from ours, since we only consider the structure based key-points, while the saliency detectors focus on visual contents that attract human attention.

### 3.3. Strength measure ($SM$) of a key-point

Some applications need a limited number of key-points such as top hundred or top thousand points. In such cases, a metric to measure the strength of the detected key-points is necessary, since many key-points given in Section 3.2 have the same level of complexity, this measure cannot be used to sort the points. Therefore, we propose a strength measure for the detector, which is conceptually similar to the corner response in many corner detectors. After thorough experiments, the following metric currently generates the best results in terms of repeatability and matching score:

$$SM = a_1 \|\boldsymbol{\alpha}\|_{l0} + a_2 \|\boldsymbol{\alpha}\|_{l1} \quad (3)$$

where $a_1, a_2$ are any positive parameters, which are selected as 1 in our experiments. The proposed metric satisfies the three following conditions to measure the strength of a key-point given by our detector:

1. A key-point is stronger if more non-zero components are required for the construction of the block around it.
2. A key-point is stronger if larger summation of quantities (coefficients) of each component is required for the construction of the block around it.
3. The metric should provide chance for lower $CM$ key-points to be stronger than higher $CM$ key-points. This condition is to diversify the types of the detected key-points, which means not only key-points with top $CM$ values, but also those with lower $CM$s could be detected even though only a limited number of key-points are expected.

### 3.4. Non-maxima suppression

After calculating the $SM$s of all detected key-points, a small window is slid through all key-points in the image. At any position, a key-point is kept if its $SM$ is higher than all of the $SM$s of its neighborhood key-points. Otherwise, it is suppressed. The suppression process contributes to the spreading of the detected key-point types, since it diminishes many nearby key-points which have very similar $CM$s. The suppression also helps improve the distribution of key-points in the image, which is good for some situations such as occlusion in which points around a neighborhood are blocked by other objects and unable to be detected by the detector.

## 4. EXPERIMENTAL RESULTS

### 4.1. Experimental setup

Currently, the sparse coding step is implemented using SPAMs library [14, 15] while other parts are implemented solely with Matlab. The dictionary used is Haar provided by Matlab. The block size is 11 by 11. The two benchmark Webcam and EF datasets [1, 2] are used for the evaluation. They are challenging datasets since the images are captured at different scenes in different contexts such as drastic changes in time, seasons, illumination, or background clutter.

Two metrics for the evaluation are repeatability and matching scores [10]. The evaluation code is provided by Mikolajczyk et al. and VLBenchmarks [10, 18]. As defined in [10], the repeatability is the ratio between the number of corresponding feature regions and the smaller number of regions in the common part of an image pair. With a known transformation between two images, two feature regions are considered corresponding if the overlap error of one region in an image and the projected region from another image is less than 0.4. The matching score, meanwhile, is the ratio between the number of correct matches and the smaller detected region number in the image pair. If the distance of a corresponding pair is the minimum in the descriptor space, the match is considered a correct one. Since a large number of features may lead to random matching of features and unreasonably high performance, 1000 key-points per image are considered as in [1, 9].

### 4.2. Quantitative results

As shown in Table 1, we compare our average repeatability with hand-crafted features such as SIFT [5], SURF [6], MSER [7], SFOP [8], Hessian Laplace [19], Harris Laplace [19], Hessian Affine [19], Harris Affine [19]. Except for our method, the performances of other detectors are collected from [9]. Since we evaluate our method with the same benchmarks, the same evaluation methods, and the same number of key-points (1000/per image), it is a fair comparison. As demonstrated in Table 1, our detector outperforms all hand-crafted detectors by a large margin in Webcam dataset. For EF dataset, it is higher than the top hand-crafted feature by 0.2%.

For the matching score, we compare our method with SIFT [5], Hessian Affine (HesAff) [19], and some state-of-the-art learning based features such as T-P24 (the best version of TIDLE) [1], Covariant point detector (CovDet) [20], and state-of-the-art detector [9]. The same evaluation procedure with SIFT descriptor is used, so the data is again collected from [9]. Table 2 shows that our detector outperforms even the top learning based detector on both datasets and has the highest average score. Our score is, however, lower than SIFT on EF. One probable reason is that unlike others, our detector is not currently implemented with the scale invariant algorithm. EF, in fact, contains drastic changes in scale, and the selection of a correct pyramid level affects the descriptor values in the evaluation.

The general high performance of our detector could be explained as follows:

1. Since all types of structures such as corners, blobs, junctions or even more sophisticated structures could be detected (See Section 5.2), we have the overall benefits of all different distinctive structures that a detector for a single type structure does not have.
2. Since all low quality structures are filtered out, these structures have no chance to decrease the performance. Other detectors for detecting a specific structure may include high and low quality structures of the type, since other structures cannot be detected to cover the lack of quality.

**Table 1. Repeatability (%) of different methods on the two benchmark datasets.**

| Method | Webcam | EF |
|---|---|---|
| SIFT | 29.5 | 20.8 |
| SURF | 46.0 | 39.7 |
| MSER | 45.1 | 37.1 |
| SFOP | 43.8 | 36.1 |
| Hessian Laplace | 51.1 | 38.8 |
| Harris Laplace | 48.2 | 35.7 |
| Hessian Affine | 42.5 | 26.6 |
| Harris Affine | 38.4 | 22.7 |
| **Ours** | **60.4** | **39.9** |

**Table 2. Matching score (%) of different methods on the two benchmark datasets.**

| Dataset | Detector | | | | | |
|---|---|---|---|---|---|---|
| | SIFT | HesAff | T-P24 | CovDet | [9] | **Ours** |
| Webcam | 12.9 | 13.8 | 13.4 | 12.0 | 19.4 | **20.4** |
| EF | 10.2 | 5.4 | 5.2 | 4.8 | 6.2 | **6.5** |
| Average | 11.6 | 9.6 | 9.3 | 8.4 | 12.8 | **13.5** |

## 5. DISCUSSION

### 5.1. The selection of sparse coding technique for the key-point detector

The purpose of sparse coding is to represent an input block with the least number of the components. The ideal case is one or zero component for each block. However, the number of components required to represent each block often varies and is larger for more complex blocks even under the pressure of the sparse coding algorithm. These complex blocks are considered distinctive blocks and good key-points for the detector. This point, in fact, separates our work from other detectors, since we consider the distinctiveness of a key-point based on the number of primitive components for the construction of the block around it, while state-of-the-art detectors consider the distinctiveness as the distinctive shapes existing in the block.

### 5.2. The capacity of detecting all types of structures

In basic linear algebra, any vector $X \in R^{N \times 1}$ can be represented as a linear combination of a complete basis set of vectors of the same dimensions. Since the dictionary we used in the sparse coding step is a complete or over-complete dictionary, whether a block $X$ represents a corner, a blob, a junction, or any other sophisticated structure, it can be decomposed into a linear combination of our basis set. If the number of nonzero components in the combination satisfies a predefined complexity range, it will be detected as a key-point by our method. To illustrate the capacity of detecting all types of structures, Figure 1 is plotted. The x-axis shows the relative $CM$ compared to the maximum $CM$ available in Figure 2. The y-axis shows the percentage of key-points of other methods that can also be detected by our method. By decreasing $CM$ from 100% to 0%, our method gradually detects all the points given by other methods.

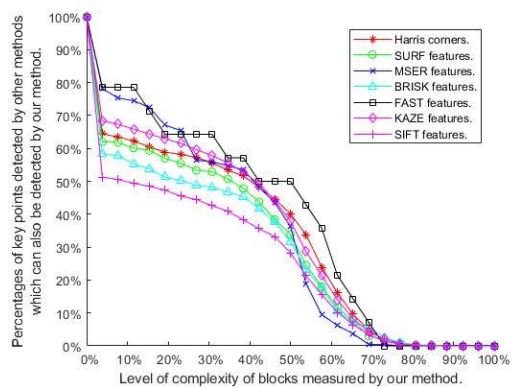

**Figure 1. An illustration for the capacity of detection of all types of structures.**

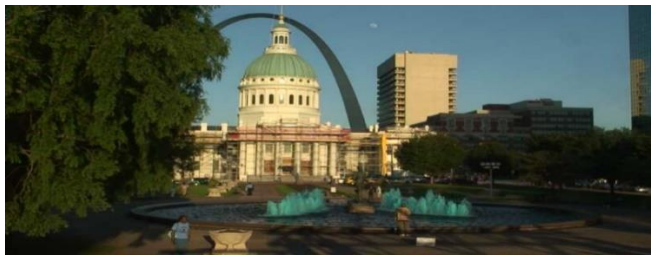

**Figure 2. An image in Webcam dataset (The image was resized for displaying).**

## 6. CONCLUSIONS

We have developed a new type of key-point detector based on sparse coding which requires no particular human designed structures to detect distinctive points in images. Experimental results have shown that our key-point detector demonstrates excellent results on two benchmark datasets. The current work opens a new direction for the researchers in the key-point detector field, and can potentially be applied in numerous vision based applications.

**We have experimentally verified that SCK detector is very robust against illumination changes. The following proof supports this claim (The proof is a part of a manuscript submitted to an IEEE Transactions in November 2018 for review and potential publication).**

**Proof:**

As presented in the ICIP paper, the input for each block $X$ to the sparse coding algorithm is its normalized version $X'_{normalized}$ after being reshaped to a vector $X'$. The formula for the normalization step is as below:

$$X'_{normalized} = \frac{X' - \mu_X}{\|X' - \mu_X\|} \quad (1)$$

where $\mu_X$ is the intensity mean of the corresponding block.

Given the following two assumptions (affine intensity change model):

1. Under change in illumination, each pixel intensity $I$ becomes $aI + b$ ($a$: multiplicative, and $b$: additional effects of the change).
2. The effects of illumination change are uniform in a small neighborhood.

We observe that under different illumination conditions, a block $X$ may be changed to $aX + b$, yet the input for the block to the sparse coding algorithm remains unchanged.

$$X'_{normalized\ after\ change} = \frac{(aX'+b)-(a\mu_X+b)}{\|(aX'+b)-(a\mu_X+b)\|} = \frac{X'-\mu_X}{\|X'-\mu_X\|} = X'_{normalized} \quad (2)$$

Thus, <u>**CM** and **SM** are also unaffected under affine intensity change</u>. In other words, the illumination invariance property of the detector is proved.